\documentclass{article}  % Comment this line out if you need a4paper
\usepackage[margin=1.0in]{geometry}
\usepackage{amsmath}
\usepackage{amssymb}
\usepackage{subcaption}
\usepackage{graphicx}
\usepackage{url}

\usepackage{makecell}
\usepackage{multirow}
\newcommand{\ie}{\textit{i.e.}, }
\newcommand{\eg}{\textit{e.g.}, }
\newcommand{\etal}{\textit{et al.}}

\title{\LARGE \bf
A Generative Approach for Detection-driven\\ Underwater Image Enhancement

}
\date{}

\author{Chelsey Edge$^{1}$, Md Jahidul Islam$^{2}$, Christopher Morse$^{3}$\\ and Junaed Sattar$^{4}$% <-this % stops a space
\thanks{The authors are with the Department of Computer Science and Engineering (CSE), Minnesota Robotics Institute (MnRI), University of Minnesota Twin Cities, Minneapolis, MN, USA.  
{\small\{$^{1}${\tt edge0037},$^{2}${\tt islam034}, $^{3}${\tt morse164}, $^{4}${\tt junaed}\}{\tt @umn.edu}}}
\thanks{*This work was supported by the US National Science Foundation Award IIS \#1845364, the University of Minnesota Doctoral Dissertation Fellowship, and the MnRI Seed Grant}% <-this % stops a space
}

\begin{document}

\maketitle
% \thispagestyle{empty}
% \pagestyle{empty}

%%%%%%%%%%%%%%%%%%%%%%%%%%%%%%%%%%%%%%%%%%%%%%%%%%%%%%%%%%%%%%%%%%%%%%%%%%%%%%%%
\begin{abstract}

In this paper, we introduce a generative model for image enhancement specifically for improving diver detection in the underwater domain. In particular, we present a model that integrates generative adversarial network (GAN)-based image enhancement with the diver detection task. Our proposed approach restructures the GAN objective function to include information from a pre-trained diver detector with the goal to generate images which would enhance the accuracy of the detector in adverse visual conditions. By incorporating the detector output into both the generator and discriminator networks, our model is able to focus on enhancing images beyond aesthetic qualities and specifically to improve robotic detection of scuba divers. We train our network on a large dataset of scuba divers, using a state-of-the-art diver detector, and demonstrate its utility on images collected from oceanic explorations of human-robot teams. Experimental evaluations demonstrate that our approach significantly improves diver detection performance over raw, unenhanced images, and even outperforms detection performance on the output of state-of-the-art underwater image enhancement algorithms. Finally, we demonstrate the inference performance of our network on embedded devices to highlight the feasibility of operating on board mobile robotic platforms.

\end{abstract}

%%%%%%%%%%%%%%%%%%%%%%%%%%%%%%%%%%%%%%%%%%%%%%%%%%%%%%%%%%%%%%%%%%%%%%%%%%%%%%%%
\section{INTRODUCTION}
\begin{figure}
\vspace{1mm}
    \centering
        \includegraphics[width=.90\linewidth]{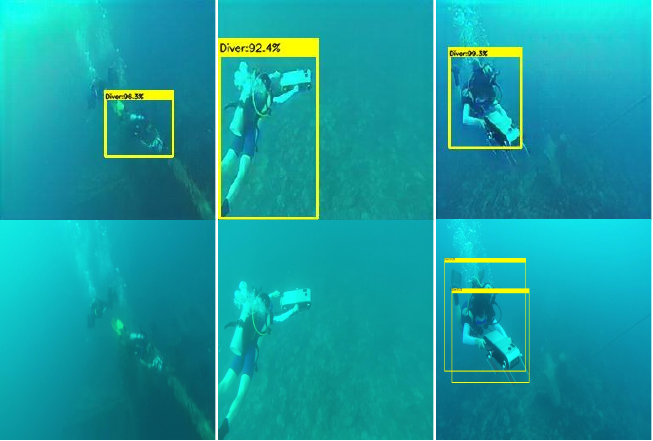}%
        \vspace{-1mm}
        \caption{Samples of detector-driven, enhanced images. The bottom images are from unenhanced ocean footage~\cite{deLangis2020VDD_C}. The top images are enhanced using our proposed model.}%
    \label{fig:detection sample}
    \vspace{-1mm}
\end{figure}
Autonomous Underwater Vehicles (AUVs) are seeing increasing use in assisting humans with tasks in underwater environments, \eg in environmental~\cite{ModasshirRobio2018_coralident} and infrastructure~\cite{petillot_pipelone} (\eg pipelines and cables) monitoring, marine archaeology~\cite{foley_archaeology,bingham_archaeology}, and underwater search and rescue. Visual perception is one of the preferred tools for many of these tasks due to the availability of cameras and their passive, energy-efficient sensing capabilities. However, due to environmental factors such as turbidity and image distortions caused by light absorption, refraction, and attenuation, the utility of visual sensing can be severely hampered. Recent work has thus looked into underwater image enhancement techniques for marine robotics and have yielded promising results. Enhancement techniques can be broadly classified into two categories: those taking the \textit{classical} approach, and those using deep machine learning algorithms.  The latter methods have seen rapid rise recently, using both Convolutional Neural Networks (CNNs)~\cite{goodfellow2016deep} and Generative models, \eg Generative Adversarial Networks (GANs)~\cite{goodfellow2014generative} and Variational Autoencoders (VAEs)~\cite{kingma2013auto}. However, the goal of these methods is usually to either alter the image to match in-air images or improve general image quality, as determined by a human viewer, focusing solely on image aesthetics. With regard to task completion, if the enhancement model is able to be executed on the AUV, these enhanced images can then be used in conjunction with visual detection or tracking models needed to complete the given task. As these tasks are often performed with a team of divers, the ability to detect, track, and follow divers will be a fundamental requirement in many applications.
 
For underwater diver tracking, traditional methods and more recent tracking-by-detection approaches have been successfully used. Islam et al.~\cite{Islam2017ICRA-MDPM}, extending the Fourier Tracker algorithm~\cite{sattar_underwater_fourier}, uses frequency-domain motion signatures of divers as a `filtering' criteria to identify tracks extracted using Hidden Markov Models~\cite{rabiner1989tutorial}. Current methods of visual diver detection and tracking include training state-of-the-art (SOTA) detection models on diver datasets (\eg \cite{deLangis2020VDD_C}) using CNN-based models \cite{islam_diver} and tracking algorithms that take into account diver motion (\eg \cite{chavez_randomforest, vehicle_fourier}).

Due to environmental factors (see Figure~\ref{fig:detection sample}) there will be situations where the trained detector does not work adequately without preprocessing or re-training. In these cases, it is shown that first using an image enhancement model and then performing diver detection can improve results~\cite{ugan, Han2020UnderwaterIP}. We take this idea a step further and propose that a \textbf{generative model designed to incorporate task requirements during the training stage can improve task-based perception}.

In other words, rather than improving images guided by human assessment, they should be enhanced based on the needs of a specific robot task. In this work, we explore image enhancement guided by the task of detecting scuba divers.
Methods directly integrating perception into GAN image enhancement have been used  for improving small-object detection~\cite{Huang_2018_small,li_perceptual_2017,liu_generative_2019,zhang_multi-task_2020}, and reducing blurriness of in-air images~\cite{prakash_it_2019}; we hypothesize that a similar approach could also improve diver detection.

In this paper, we propose a \textbf{D}etection-driven \textbf{Un}derwater \textbf{I}mage \textbf{E}nhancement model based on \textbf{G}enerative \textbf{A}dversarial \textbf{N}etworks, which we refer to as \textbf{DUnIE-GAN}. In addition to focusing on learning image-to-image translation from a \textit{distorted} to \textit{enhanced} domain, the proposed model makes use of bounding box locations and classification information provided by a pre-trained diver detection model to guide the enhancement towards better detection results. We provide insights into its design choices and present experimental results showing the positive effects of incorporating additional detection information into a GAN-based image enhancement model such as FUnIE-GAN~\cite{funie}. Moreover, we quantify the improved diver detection results for a variety of cases and perform extensive image quality analysis on the DUnIE-GAN-enhanced images. We also conduct performance comparison with several SOTA underwater image enhancement models that further validates the effectiveness of our proposed approach.

\section{Background}

\subsection{Visual Diver Detection}
Visual detection and following of human divers by autonomous robots is well-studied for its usefulness in human-robot cooperative underwater missions~\cite{sattar2008enabling,gomez2019caddy}.
Due to operational simplicity and computational efficiency, the simple feature-based detectors followed by standard model-free trackers
have been traditionally used in autonomous diver-following applications~\cite{sattar2006performance,islam2019person}. 
Particle filters and optical flow-based methods are also utilized for tracking divers in spatio-temporal domain~\cite{islam2019person}. Since color distortions and low visibility issues are common underwater, the frequency-domain cues~\cite{islam2017mixed,sattar2009underwater} of divers' motion are used for reliable detection as well. Besides, several feature-based learning approaches~\cite{islam2019person,sattar2009robust} such as SVMs and ensemble methods have been investigated for diver tracking and underwater object tracking in general. However, these methods lack generalizability and often fail in noisy visual conditions.

To address the inherent difficulties of underwater vision, several CNN-based diver detection models are proposed recently~\cite{islam_diver,de2020realtime,islam2018understanding}. Once trained with large-scale comprehensive data, these models can achieve invariance to the appearance, scale, or orientation of divers (to the camera). They also provide a considerably better generalization performance than classical approaches. Despite the high learning capacity, in practice, their performance gets severely affected by marine artifacts such as noise and optical distortions, which result in low-contrast, often blurred, and color-degraded images~\cite{funie}. Various image enhancement techniques are generally adopted to mitigate these challenges, \ie to restore the perceptual and statistical qualities of the input images before passing it to the diver detection pipeline.

\subsection{Enhancing Underwater Imagery}
The underwater image enhancement problem deals with correcting non-linear image distortions caused by the particularities of light propagation underwater~\cite{bryson2016true,akkaynak2018revised}. 
Some of these aspects can be modeled and well estimated by physics-based solutions by exploiting prior knowledge (\eg haze-lines and dark channel prior~\cite{bryson2016true,berman2018underwater}) or making statistical assumptions (\eg adopting an atmospheric dehazing model~\cite{he2010single,Perez_dehaze}). However, these approaches require scene depth and water-quality measures for accurate modeling, which are not always available in practice.

As a practical alternative, learning-based approaches have been widely explored and they have demonstrated inspiring success in recent years. Several models based on deep CNNs and GANs report SOTA performance~\cite{islam2020sesr,funie,Li_waternet} on benchmark datasets. Driven by large-scale supervised training, these approaches learn sequences of non-linear filters to approximate the underlying pixel-to-pixel mapping~\cite{Isola2017ImagetoImageTW} between the \textit{distorted} and \textit{target} image domains. 
The contemporary CNN-based generative and residual networks (\eg Deep SESR~\cite{islam2020sesr}, WaterNet~\cite{Li_waternet}) are shown to be very effective in learning such mapping.
Moreover, the GAN-based models (\eg FUnIE-GAN~\cite{funie}, UGAN~\cite{ugan}, Fusion-GAN~\cite{li2019fusion}) attempt to improve generalization performance by employing a two-player min-max game, where an adversarial \textit{discriminator} evaluates the \textit{generator}-enhanced images compared to ground truth samples. This forces the generator to learn realistic enhancement while evolving with the discriminator toward equilibrium. While the existing models provide good generic solutions for perceptual enhancement, learning to recover image distortions for specific tasks such as diver detection or classification has not been explored in-depth in the literature.

\subsection{GANs and Object Detection}
While not extensively explored for underwater applications, integrating object detection components within the GAN architecture has been well-studied in the terrestrial domain. For instance, the auxiliary classifier
GANs~\cite{cgan} are proposed to synthetically generate realistic images, while several variants of the Multi-task GAN (MT-GAN)~\cite{zhang_multi-task_2020} and Perceptual GAN~\cite{li_perceptual_2017} are used to improve small-object detection performance by using image super-resolution techniques. In such models, the components of bounding box regression and classification are determined through discriminator branches using additional fully-connected layers. 
Besides, Liu \etal~\cite{liu_generative_2019} integrate a RetinaNet~\cite{retinanet} detector in DetectorGAN; however, this is done for generating data to train an object detector rather than improving its performance through image enhancement. 

Our model differs firstly in that we formulate a detection-guided style transfer 
problem to learn to enhance the distorted images rather than performing enhancement or super-resolution separately. Secondly, our model attempts to enhance images for use in conjunction with a pre-trained object detector for specific task execution.

\section{Methodology}
\begin{figure*}
\vspace{2mm}
    \centering
        \includegraphics[width=.70\linewidth]{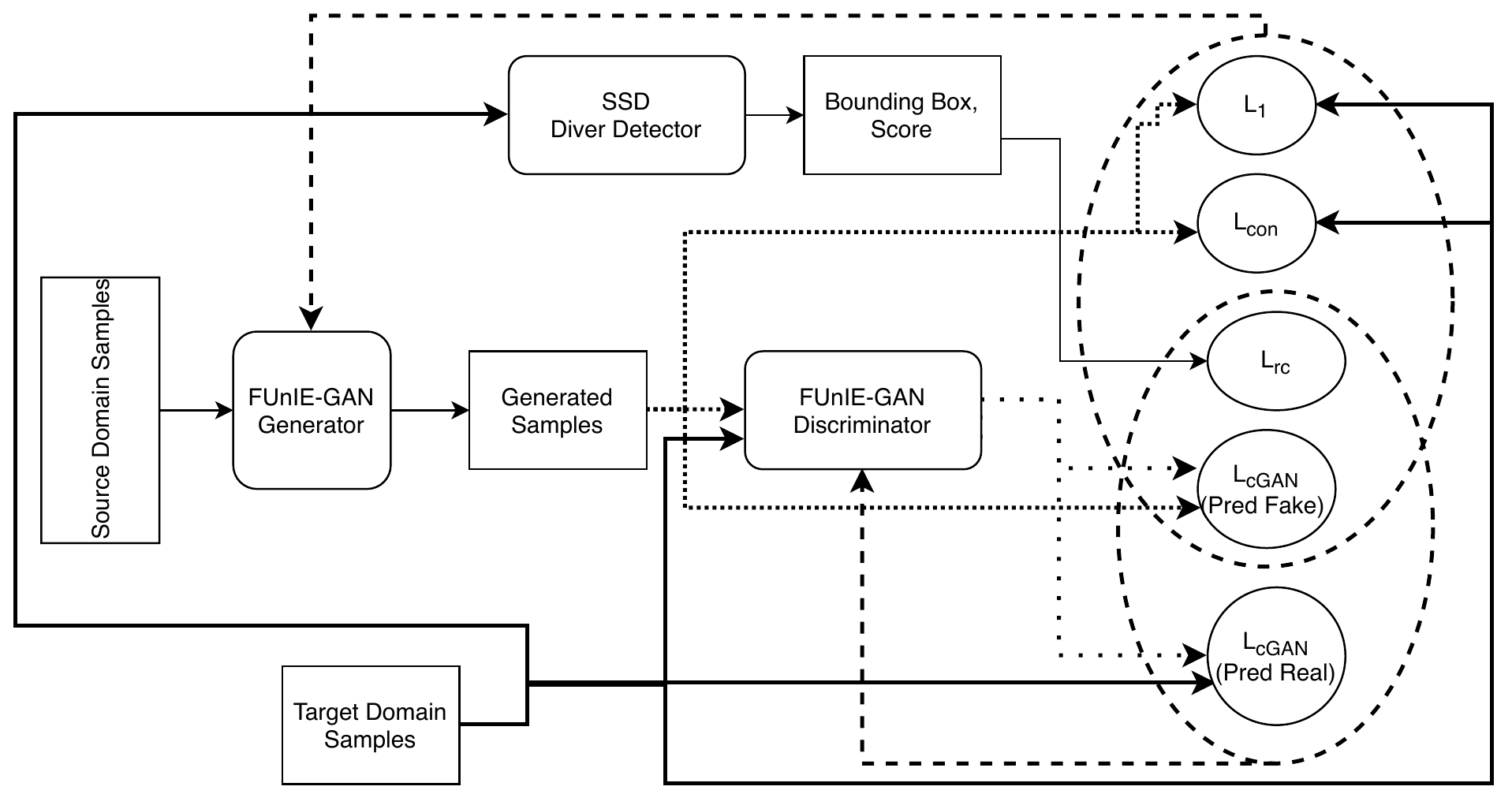}%
        %\vspace{-1mm}
        \caption{Proposed Model Architecture: Of modular design, the generator and discriminator incorporate FUnIE-GAN\cite{funie} networks. The diver detector can be replaced with any trained detector.}%
    \label{fig:architecture}
    \vspace{-1mm}
\end{figure*}

As shown in Figure~\ref{fig:architecture}, our proposed DUnIE-GAN model integrates a pre-trained diver detector with a GAN-based image enhancement model. Similar to the existing learning-based models~\cite{funie,ugan}, given a source domain $X$ of distorted underwater images and target domain $Y$, our goal is to learn a mapping $G: X \rightarrow Y$. However, instead of learning a map that perceptually enhances the resulting image, our goal is to enhance the images in a way that improves detection results of the diver detector. For the purposes of this paper, the only diver detector used is a trained Single Shot Detector (SSD)~\cite{liu2016ssd} model with a MobileNet V2 backbone\footnote{The model weights are imported from \url{https://github.com/IRVLab/deep-diver-following}.}.%, which we have ported to Python3.

\begin{figure}
\vspace{2mm}
    \centering
        \includegraphics[width=.95\linewidth]{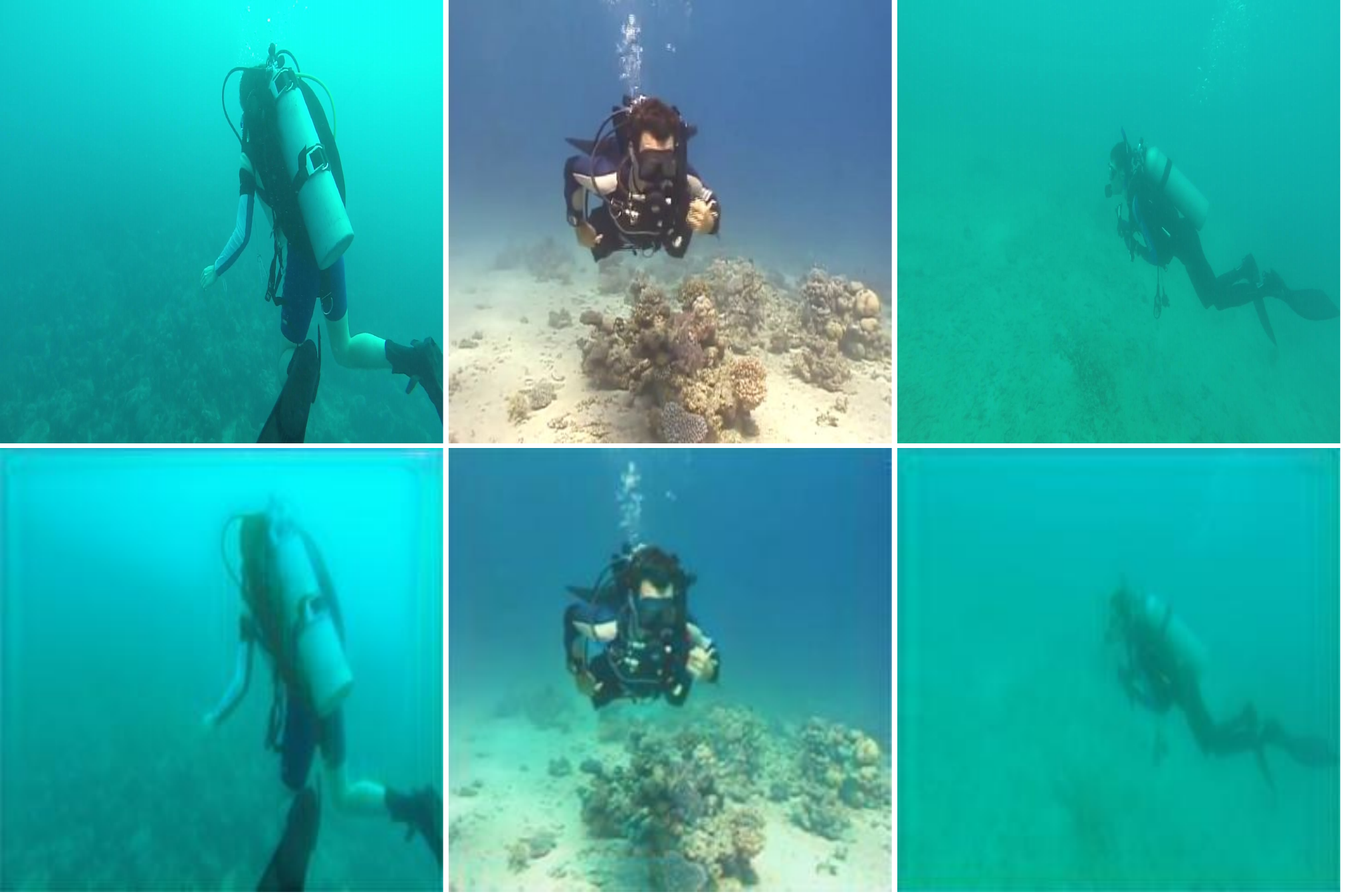}%
        \vspace{-1mm}
        \caption{A few sample training pairs are shown; images on the top and bottom row belong to the \textit{target domain} and \textit{distorted domain}, respectively.}%
    \label{fig:trainingpairs}
    \vspace{-1mm}
\end{figure}

\subsection{Data preparation}
For the supervised training, our dataset was prepared with the end goal of better diver detection in mind. Our dataset was created from the combined ocean data from~\cite{islam_diver} and~\cite{deLangis2020VDD_C}, which contain single diver images with a variety of environmental factors. We then used the diver detector to select images in which the diver was detected with a $0.55$ accuracy score to ensure possible (detector determined) true positive detections during the training process. These images were then distorted based on the method used by Islam~\etal~\cite{funie} in order to create a paired dataset; see Figure~\ref{fig:trainingpairs} for some training image samples. Each training image includes only one diver. Hand-labeled annotations of these images are used for the ground truth bounding box data.

\subsection{Overarching Model}
Following the standard paired training pipelines of \cite{ugan,funie}, we use distorted images as input to the generator. The generator then produces synthetic images, which are sent to the discriminator to get its validity response. As mentioned earlier, the generator and discriminator then play a min-max game to eventually learn the mapping $G: X \rightarrow Y$. In our model, we also use the synthetic images as input for the diver detector to find detector-specific bounding boxes and classification scores, which are used in an additional detection-based loss function.

\subsection{Generative Adversarial Network Architectures}
The GAN portion of our model is based on the FUnIE-GAN~\cite{funie} architecture. The generator network follows the U-Net~\cite{Ronneberger2015UNetCN}-like encoder-decoder structure with an input resolution of $256\times256\times3$; the encoder eventually learns $256$ feature-maps of size $8\times8$. The decoder uses these feature-maps to learn to generate an enhanced $256\times256\times3$ image. The discriminator uses the Markovian Patch-GAN~\cite{Isola2017ImagetoImageTW} architecture, which takes as input the combination of real and generated images in $256\times256\times6$ form and outputs a $16\times16\times1$ real or fake response. Specific network details on FUnIE-GAN architecture can be found in~\cite{funie}.  As the generator and discriminator networks are not altered, we are able to take advantage of seeding our model with a pre-trained FUnIE-GAN model to initiate training.

\subsection{Diver Detector}
Our model incorporates the previously mentioned SSD (MobileNet V2) pre-trained diver detector. We incorporate this model outside of the GAN architecture, with the detector being independent from the generator and discriminator networks. This design is intentional in order to create a modular network that could be used with other types of detectors or image enhancement models. For the training process, we set the detection threshold to $0.01$ and choose to only return the detected bounding box with the highest classification score to tune DUnIE-GAN towards improving detections after enhancement.
For each batch during training, the detector runs on every generated image and outputs a bounding box $b = (x_{min}, y_{min}, w, h)$, as well as the confidence score $s \in [0,1]$; here, $(x_{min}, y_{min})$ is an image coordinate, whereas $w$ and $h$ represent width and height of the box, respectively. 
These outputs are subsequently used in the detection loss $\mathcal{L}_{rc}$ described below.

\subsection{Objective Function}
Like the learning pipeline of FUnIE-GAN~\cite{funie}, our objective function incorporates global similarity and image content losses, as well as the conditional adversarial loss. Additionally, we design a detection loss similar to those used in~\cite{Huang_2018_small} and~\cite{zhang_multi-task_2020} which includes bounding box regression and classification portions.

\textbf{Base GAN Loss}: We use the standard conditional adversarial loss~\cite{DBLP:MirzaO14_conditional} to learn the mapping of $G: \{X,Z\} \rightarrow Y$ where $X(Y)$ represents the distorted (target) domain and $Z$ represents random noise; it is defined as follows 
\begin{equation} \label{eq:cgan}
\centering
  \begin{aligned}
    \mathcal{L}_{cGAN}(G,D) &= \mathbb{E}_{X,Y} \big[\log D(Y)\big]  \\
     & + \mathbb{E}_{X,Y} \big[\log (1-D(X, G(X, Z)))\big].
  \end{aligned}
\end{equation}

Since we also want to maintain the same aspects between the target and generated images, as in \cite{funie}, we also include the loss terms for global similarity ($\mathcal{L}_{1}$) and image content ($\mathcal{L}_{con}$)  as follows:
\begin{equation} \label{eq:similarity}
\centering
  \begin{aligned}
    \mathcal{L}_{1}(G) &= \mathbb{E}_{X,Y,Z} \big[\| Y-G(X,Z) \|_{1}\big],  
  \end{aligned}
\end{equation}
\begin{equation} \label{eq:content}
\centering
  \begin{aligned}
    \mathcal{L}_{con}(G) &= \mathbb{E}_{X,Y,Z} \big[\| \Phi(Y)-\Phi(G(X,Z))\|_{2}\big].  
  \end{aligned}
\end{equation}

\textbf{RC Loss}: Here, we include the output of the diver detector into the objective. Let $s$ be the confidence score, and $b_{d}$ and $b_{a}$ be the bounding box information from the detection and ground truth annotations, respectively. Our detection loss is then calculated as
\begin{equation} \label{eq:percept}
\centering
  \begin{aligned}
\mathcal{L}_{rc} = \mathcal{L}_{reg}(b_{d},b_{a})+\mathcal{L}_{cls}(s),
  \end{aligned}
\end{equation}
where $\mathcal{L}_{reg}(b_{d},b_{a})$ is the following loss proposed in \cite{fastrcnn}
\begin{equation} %\label{eq:percept}
\centering
  \begin{aligned}
  Smooth_{L_{1}}(b_{d},b_{a}) =
  \begin{cases}
                                   0.5(b_{d}-b_{a})^{2} & \text{if $|b_{d}-b_{a}|<1$}, \\
                                   |b_{d}-b_{a}|-0.5 & \text{otherwise;}
  \end{cases}
  \end{aligned}
\end{equation}
and $\mathcal{L}_{cls}=-\log s$. The use of these losses is inspired by the methods used in \cite{zhang_multi-task_2020, li_perceptual_2017} and \cite{Huang_2018_small}. If there is no detection found in a specific image, an arbitrary bounding box and score are introduced. This in turn returns a high loss value for that specific image in order to weight the model towards accurate detection.

Besides, Liu~\etal~\cite{liu_generative_2019} mentions that adding a loss similar to $\mathcal{L}_{rc}$ to the generator may minimize performance on real images as the model will be tuned towards good detections on synthetic images. However, Zhang~\etal~\cite{zhang_multi-task_2020} suggests including a similar loss in optimizing the generator to help enforce a semantic level similarity constrain. Our ablation experiments investigate the effects of using such loss terms in the learning, which we discuss in Section~\ref{loss_ablation}.

\textbf{Objective function}: Using the loss components defined in~\cite{funie} as well as the new detector-based loss formulated in Eq.~\ref{eq:percept}, our final objective function is expressed as  
\begin{equation} \label{eq:train_pipeline}
\centering
  \begin{aligned}
    G^{*}=\arg\min_{G}\max_{D}&\mathcal{L}_{cGAN}(G,D) + \lambda_{1}\mathcal{L}_{1}(G) \\
         &+\lambda_{c}\mathcal{L}_{con}(G)+\mathcal{L}_{rc}.
  \end{aligned}
\end{equation}
We use $\lambda_1 = 0.7$ and $\lambda_c = 0.3$ as in \cite{funie} as scaling factors.

\section{Experimental Results}

\subsection{Implementation details}
Our model is implemented using the PyTorch libraries\footnote{PyTorch libraries: \url{https://pytorch.org/}} for the majority of the optimization pipeline. However, the trained diver detector is implemented in Tensorflow~\cite{abadi2016tensorflow}, so our environment includes both libraries. Each DUnIE-GAN model was trained on a Linux Machine using a NVIDIA GTX 1080 GPU. 
DUnIE-GAN was trained on $8$k images with single annotated divers. Various categories of test sets are created by images from~\cite{islam_diver} and~\cite{deLangis2020VDD_C} that were excluded from the training. More detailed information on these test sets can be found in Section~\ref{test set}.

Our model uses pre-trained weights of FUnIE-GAN on the EUVP dataset\footnote{FUnIE-GAN (PyTorch) model is available at: \url{https://github.com/xahidbuffon/FUnIE-GAN}} for 100 epochs, and then it is further tuned with our training pipeline in Eq.~\ref{eq:train_pipeline}. %By using validation that includes a detection metric, 
%For the ablation experiments, 
The convergence of different training models fall between $80$-$160$ additional epochs. In particular, the model using $\mathcal{L}_{rc}$ to optimize the generator and discriminator is trained for $120$ epochs, while the model with $\mathcal{L}_{rc}$ in the discriminator is trained $140$ epochs. A batch size of $32$ is used in all the training modes; here, we use a larger batch size than is used in~\cite{funie} to increase the probability of images with detectable divers.
 
\subsection{Loss Ablation}\label{loss_ablation}
We first present an ablation study based on how the object detection loss is integrated with our model. We abbreviate the location of $\mathcal{L}_{rc}$ loss optimizations as follows:
\begin{itemize}
    \item For both generator and discriminator: DUnIE-GAN-B 
    \item For generator: DUnIE-GAN-G 
    \item For discriminator: DUnIE-GAN-D 
    \item For neither: DUnIE-GAN-N 

\end{itemize}

In particular we are interested in the results of DUnIE-GAN-B and DUnIE-GAN-D in reference to the insights gained from Liu \etal~\cite{liu_generative_2019} and Zhang \etal~\cite{zhang_multi-task_2020}. We also make note that as our base network architecture and losses are inspired by FUnIE-GAN, results of DUnIE-GAN-N would be similar to those of pre-training FUnIE-GAN with the EUVP dataset and then additionally training with our diver dataset.

As we are interested in increasing the diver detection accuracy, we use DUnIE-GAN in conjunction with the diver detector used in training our model. We use Average Precision (AP)~\cite{padillaCITE2020} and Intersection over Union (IoU) as our performance metrics. Mean IoU per image is found, with penalties of an additional score of $0$ included for each extra detection per image. The mean of all IoUs over the dataset is then taken. All tests are performed with the diver detector score threshold of $0.55$. The overall performance is determined based on $415$ distorted single diver images that are not used for training or validation. The results are illustrated in Table~\ref{tab:ablation}. The baseline for this evaluation is determined by using the diver detector directly on the distorted images. It can be seen that while all models increase detection results, DUnIE-GAN-B and DUnIE-GAN-D have the highest scores of IoU and AP, respectively. To test these results on a more generalized task, we use these two models in comparison with other underwater enhancement tasks on non-distorted images.

\begin{table}[h]
\vspace{2mm}
\centering
\caption{Ablation results for $\mathcal{L}_{rc}$: AP and mean IoU on distorted underwater images of single divers (baseline: AP = $63.38$ and IOU = $33.9$).}
%\scriptsize
%\vspace{-1mm}
\begin{tabular}{c|c|c}
  \hline
   Model & AP (\%) & IoU (\%) \\
     \hline  \hline
   DUnIE-GAN-B  & 70.08 & 44.2 \\
  \hline
   DUnIE-GAN-G & 69.34 & 43.8 \\
  \hline
   DUnIE-GAN-D & 71.26 & 43.8 \\
  \hline
   DUnIE-GAN-N & 69.78 & 43.2 \\
    \hline
 \end{tabular}
\label{tab:ablation}
\end{table}

\subsection{Ocean Image Test Set and Comparison Models} \label{test set}
We compare DUnIE-GAN with several underwater image enhancement models based on the results of the task of diver detection as well as on the results of image quality metrics. The test set, like the training set, contains images taken directly from~\cite{islam_diver} and~\cite{deLangis2020VDD_C} in various categories based on thresholding detected divers at $0.55$ with the detector. The various categories include:
\begin{itemize}
    \item Single Diver (SD)-Detected: $415$ images
    \item Single Diver (SD)-Undetected: $1014$ images 
    \item Multiple Divers (MD)-All Detected: $1021$ images
    \item Multiple Divers (MD)-Some Detected: $1025$ images
\end{itemize}
These sets are not synthetically distorted; also note that SD-Detected are the undistorted images used in creation of images for \textit{testing} DUnIE-GAN ablation and have not been seen by the model during training. The enhancement models that we use for comparison include the physics based model
multi-scale Retinex (MS-Retinex)~\cite{zhang_underwater_2017retinex} 
and the learning-based models WaterNet~\cite{Li_waternet}, Underwater GAN (UGAN)~\cite{ugan}, Fast Underwater Image Enhancement GAN (FUnIE-GAN)~\cite{funie}, and Deep Simultaneous Enhancement and Super-Resolution (Deep SESR)~\cite{islam2020sesr}. All learning-based models are trained on the EUVP~\cite{funie} dataset. 
\begin{figure}[h]
\vspace{2mm}
    \centering
        \includegraphics[width=.98\linewidth]{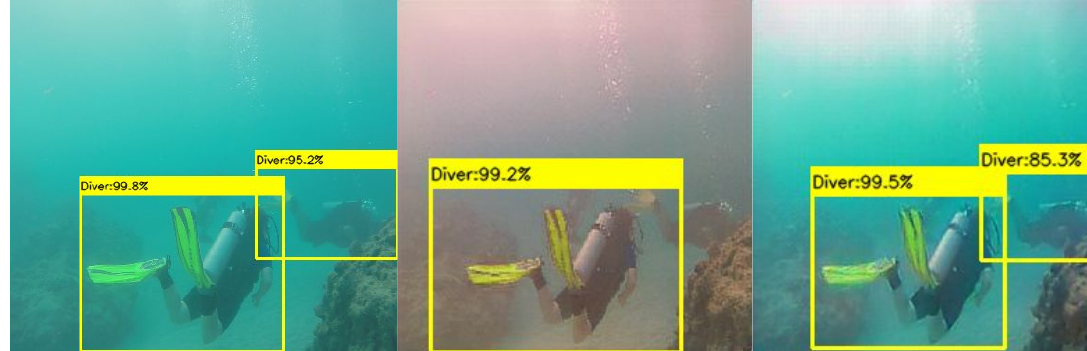}%
        %\vspace{-1mm}
        \caption{The Importance of task-based enhancement: from left to right, samples from Baseline, FUnIE-GAN, and DUnIE-GAN are shown. Although FUnIE-GAN-enhanced images score higher on UIQM metrics and the images look perceptually enhanced, this does not guarantee all divers will be detected. The bounding box locations are also more accurate and concise in this DUnIE-GAN-generated sample.}%
    \label{fig:funiesample}
    \vspace{-1mm}
\end{figure}
\subsection{Diver Detection Comparisons}

For the comparison of diver detection performance, we use the same metrics included in the loss ablation: mean AP and mean IoU. We evaluate the results of the diver detector after enhancing the images using the various models. The baseline is created by running the diver detector directly on the dataset images. The results of this comparison can be seen in Table~\ref{tab:detect_compare}. We find an interesting result that on this particular dataset, none of the enhancement models return AP results better the baseline images. However, both DUnIE-GAN model results are competitive with the enhancement models, while DUnIE-GAN-B achieves the highest scores on AP and IoU for the SD-Detected dataset. We posit that one reason Deep SESR  does very well on this task is the model takes into account an additional information loss for saliency predictions during its training~\cite{islam2020sesr}. 
\begin{table}[h]
\vspace{2mm}
\centering
\caption{Comparison of diver detection performance on enhanced images based on AP and mean IoU. 
}
\scriptsize
\begin{tabular}{l|l|r|r|r|r|r|r|r|r}
  \hline
    &Dataset&Baseline&MS-Retinex & WaterNet & UGAN & \makecell{FUnIE- \\ GAN} & \makecell{Deep\\ SESR} & \makecell{\textbf{DUnIE-} \\ \textbf{GAN-B}} &  \makecell{\textbf{DUnIE-} \\ \textbf{GAN-D}}\\
     \hline \hline
      \parbox[t]{1mm}{\multirow{4}{*}{\rotatebox{90}{\textbf{AP(\%)}}}}&SD-Detected& $91.86$ &$11.24$ & $56.46$ & $62.22$ & $64.32$ & $69.88$ & $75.68$&$68.73$\\
    
      &SD-Undetected&$25.56$ & $1.08$ & $7.89$ &$9.18$ & $19.84$ &$20.23$ & $18.36$ &$19.03$\\
      &MD-All Detected& $92.00$ & $2.83$ & $41.11$ & $51.66$ & $66.67$ & $82.27$ & $71.76$ & $76.34$ \\

      &MD-Some Detected& $54.45$ & $2.42$ & $21.43$ & $29.38$ & $40.78$ &$51.62$ & $44.89$ & $47.01$\\
      \hline 
      \parbox[t]{1mm}{\multirow{4}{*}{\rotatebox{90}{\textbf{IoU(\%)}}}}&SD-Detected& $67.3$ &$12.5$ & $33.0$ & $38.3$ & $42.3$ & $45.5$ & $48.2$&$44.0$\\
      &SD-Undetected&$2.8$ & $1.6$ & $2.3$ &$3.2$ & $5.0$ &$5.3$ & $9.2$ &$7.5$\\
      &MD-All Detected& $64.9$ & $4.5$ & $23.9$ & $27.5$ & $36.5$ & $49.3$ & $41.8$ & $44.2$ \\
      &MD-Some Detected& $25.4$ & $3.4$ & $12.0$ & $14.6$ & $18.2$ &$22.1$ & $21.5$ & $21.3$\\
      \hline 
\end{tabular}
\label{tab:detect_compare}
\end{table}
With regards to mean IoU, we again observe favorable results with both DUnIE-GAN models, and particularly in the case of SD-Undetected, having the highest scores. As we intentionally train the model for minimization of distance between bounding boxes, an increase in IoU when objects are detected is an encouraging result.

\subsection{Image Quality Analysis}
\begin{figure}[h]
\vspace{-1mm}
    \centering
        \includegraphics[width=.98\linewidth]{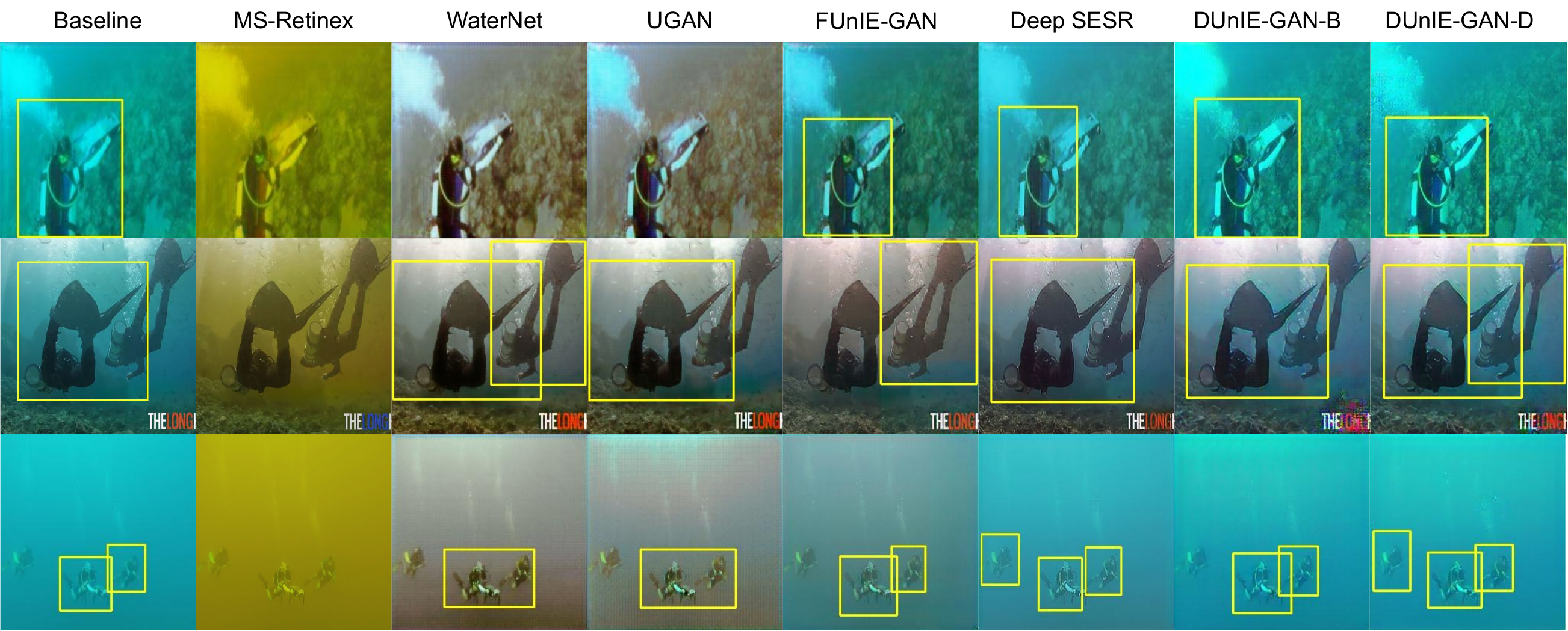}%
        \vspace{-1mm}
        \caption{Sample results of diver detections over the enhancement comparison models. All detections with a confidence threshold score over $0.55$ are labeled.}%
    \label{fig:evalsamples}
    \vspace{-3mm}
\end{figure}

To analyze the enhancement, we conduct a performance comparison of all models using the underwater image quality metric (UIQM)~\cite{7305804uiqm}. We choose this evaluation as it is inspired by properties of human vision systems and evaluates image quality in methods that are shown to relate to human preference. A higher overall score lends itself to a higher correlation with perceived image quality. Results of this comparison are found in Table~\ref{tab:uiqm}. As our goal is to show task-specific enhancement, rather than enhancement for human perception, we are not looking to have the \textit{best} score but rather looking to compare these results with the capability of diver detection. We find that while WaterNet and UGAN receive high UIQM scores, it does not translate to higher detection results. See Figure~\ref{fig:evalsamples} for a visual comparison. As our model is designed for task-specific goals, depending on the task, it may be preferable for the enhanced image to have UIQM scores similar to those of the original image.

\begin{table*}[h]
\centering
\caption{Comparison of UIQM scores on the same test sets as of Table~\ref{tab:detect_compare}; scores are shown as $mean \pm \sqrt{variance}$.}
\scriptsize
\begin{tabular}{l|c|c|c|c|c|c|c|c}
  \hline
   \textbf{Dataset}&Baseline& MS-Retinex & WaterNet & UGAN & \makecell{FUnIE- \\ GAN} & \makecell{Deep \\ SESR} & \makecell{\textbf{DUnIE} \\ \textbf{-GAN-B}} &  \makecell{\textbf{DUnIE} \\ \textbf{-GAN-D}}\\
      \hline \hline
       SD-Detected&$1.05 \pm 0.37$&$1.01 \pm0.38 $&$ 2.48\pm0.40 $&$2.58 \pm 0.36 $&$ 2.37\pm.048 $&$1.87 \pm0.42 $&$ 1.46\pm 0.34$&$1.40 \pm0.36 $\\
    SD-Undetected&$1.20 \pm 0.36$&$1.16 \pm 0.39$&$ 2.78\pm0.36 $&$2.82 \pm0.37 $&$2.57 \pm 0.44$&$ 2.05\pm 0.44$&$ 1.52\pm0.33 $&$ 1.48\pm0.33 $\\
    MD-All Detected&$0.87 \pm0.31 $&$0.83 \pm0.29 $&$2.31 \pm 0.26$&$2.44 \pm 0.25$&$ 2.15\pm 0.37$&$1.66 \pm0.31 $&$1.33 \pm0.33 $&$1.27 \pm0.32 $\\
      MD-Some Detected&$ 0.90\pm 0.37$&$ 0.86\pm0.37 $&$2.40 \pm 0.33$&$2.51 \pm 0.30$&$2.18 \pm0.46 $&$ 1.70\pm0.40 $&$1.34 \pm 0.36$&$1.28 \pm0.37 $\\
  \hline
\end{tabular}
\label{tab:uiqm}
\end{table*}

\subsection{Runtime Feasibility}
As a task-driven enhancement model, it is necessary for DUnIE-GAN to be able to run in real-time on an AUV or ROV. As the architecture is similar to that of FUnIE-GAN, it is expected that the operation rate will be similar, when using the Pytorch version. DUnIE-GAN runs at a rate of $113.15$ FPS (frames per second) on a NVIDIA Jetson Xavier and $838.8$ FPS on a NVIDIA RTX-2080. FUnIE-GAN, Pytorch, runs at $119.205$ FPS and $887.1$ FPS, respectively. For this reason, we claim that our model is capable of use onboard an AUV or ROV.

\section{Conclusion}

In this paper, we introduce the objective of task-based image enhancement for the underwater domain. We propose a model inspired by current image enhancement techniques, and create a GAN-based architecture for image enhancement for the purpose of diver detection, which can be used on an AUV or ROV. We also perform an ablation test to investigate the use of different objective functions. While the model is trained on synthetic images, we show generalization to raw ocean images. The results of a diver detection model used in conjunction with various image enhancement models are compared with our model. We show that images enhanced for the human perspective do not always lead to the best results for a task assigned to a robotic platform. In the future, we plan to improve upon this idea through an exploration of other tasks in addition to detecting divers through bounding boxes, such as for image segmentation. We also intend to explore the introduction of other objective functions, such as incorporating information loss to predict saliency.

\clearpage
\bibliographystyle{IEEEtran}
\bibliography{bibliograph}

\end{document}